\documentclass[runningheads]{llncs}
\usepackage[T1]{fontenc}
\usepackage{hyperref}
\usepackage{amsmath}
\usepackage{multirow}
\usepackage{graphicx,verbatim}
\usepackage{amssymb}

\begin{document}
\title{Fine-Grained Rib Fracture Diagnosis with Hyperbolic Embeddings: A Detailed Annotation Framework and Multi-Label Classification Model}
%
\author{Shripad Pate\inst{1} \and
Aiman Farooq\inst{1} \and
Dr. Suvrankar Datta\inst{2} \and
Dr. Musadiq Aadil Sheikh\inst{3} \and
Dr. Atin Kumar\inst{2} \and
Deepak Mishra\inst{1}}
\authorrunning{F. Author et al.}
%
\institute{Indian Institute of Technology Jodhpur, 342030 India \\
\email{farooq1.@iitj.ac.in}\\
 \and
All India Institute of Medical Sciences Delhi, 110029\\
\and
Government Medical College Srinagar, 190010
}


\maketitle              
\begin{abstract}
Accurate rib fracture identification and classification are essential for treatment planning. However, existing datasets often lack fine-grained annotations, particularly regarding rib fracture characterization, type, and precise anatomical location on individual ribs. To address this, we introduce a novel rib fracture annotation protocol tailored for fracture classification. Further, we enhance fracture classification by leveraging cross-modal embeddings that bridge radiological images and clinical descriptions. Our approach employs hyperbolic embeddings to capture the hierarchical nature of fracture, mapping visual features and textual descriptions into a shared non-Euclidean manifold. This framework enables more nuanced similarity computations between imaging characteristics and clinical descriptions, accounting for the inherent hierarchical relationships in fracture taxonomy. Experimental results demonstrate that our approach outperforms existing methods across multiple classification tasks, with average recall improvements of 6\% on the AirRib dataset and 17.5\% on the public RibFrac dataset.


\end{abstract}
\section{Introduction}
\begin{figure*}[!t] \centering \includegraphics[width=\textwidth,height=2cm]{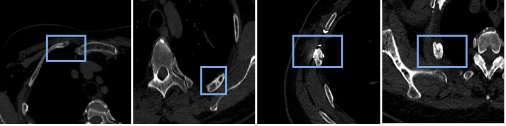} \caption{CT scans showing different types of rib fractures: (a) Buckle fracture in the anterior Right rib, (b) Non-displaced fracture in the posterior left rib, (c) Displaced oblique fracture in the lateral right rib, and (d) Severely displaced fracture.} \label{fig:sample} \end{figure*}

The automated recognition of rib fractures in CT scans is critical for emergency trauma assessment, follow-up care, and treatment planning ~\cite{franssen2024treatment}. Identification of 
rib fractures present unique diagnostic challenges due to their complex anatomical characteristics. Fractures manifest in various morphological patterns, from simple buckle fractures to complex comminuted fractures with varying degrees of displacement. As shown in Figure~\ref{fig:sample} with each rib following an oblique course across multiple CT slices ~\cite{ringl2015ribs}. Recognizing these patterns have distinct clinical implications that directly influence treatment decisions. The 6-point ``RibScore'' system~\cite{chapman2016ribscore}, which identifies radiographic parameters linked to complications: (A) six or more fractures, (B) bilateral fractures, (C) flail chest, (D) three or more severely displaced fractures, (E) first rib fracture, and (F) fractures across all three anatomical areas. However, most current public datasets~\cite{jin2020deep} are largely limited to binary fracture detection, lacking annotations for morphological subtypes and clinical severity markers.

Deep learning (DL) models have shown great promise for rib fracture detection. Early approaches, like FracNet ~\cite{jin2020deep}, built on 3D-UNet ~\cite{ronneberger2015u} with sliding window sampling to reduce complexity in non-rib regions. Recent developments include specialized architectures such as Liu et al.'s ~\cite{liu2021multi} multi-scale segmentation network, Wu et al.'s \cite{wu2021development} F-RCNN with ResNet50 for rib-level diagnosis, and Li et al.'s ~\cite{li2023automatic} CenterNet+DLA-34 for ROI detection. To address fracture analysis complexity,~\cite{zhou2021automatic} integrated imaging and clinical data through multi-modal fusion, ~\cite{cao2023robust} introduced SA-FracNet with contrastive learning on unlabeled CT data. Castro-Zunti et al. ~\cite{castro2024ribfracturesys} developed a multi-class classifier to identify an unfractured, a newly fractured, a previously fractured, and now healed rib. They also track the rib number using CV algorithm. ~\cite{yang2024deep} developed a detection and classification model that characterizes fracture into buckle, non-displaced, displaced, or segmental.
However, these methods mainly focus on binary detection or simple classification tasks, failing to address the challenge of simultaneously identifying multiple fracture characteristics and patterns.

Recently, hyperbolic geometry has emerged as a powerful framework for modeling hierarchical relationships in DL. Khrulkov et al. ~\cite{khrulkov2020hyperbolic} showed hyperbolic embeddings improve image classification by better capturing visual hierarchies, while Desai et al. ~\cite{desai2023hyperbolic} extended this to vision-language models. Despite medical imaging's inherently hierarchical nature, hyperbolic geometry remains unexplored. In this context, we propose a novel hyperbolic learning framework for comprehensive rib fracture analysis. This paper makes three main contributions: 
\begin{itemize}

\item We establish a fine-grained annotation protocol for rib fractures, where each fracture is labeled with key clinical attributes, including anatomical location, displacement severity, morphological patterns, and its contribution to segmental or flail chest injuries.

\item We develop a two-stage pipeline for fine-grained rib fracture analysis. The first stage employs a modified Faster R-CNN~\cite{ren2016faster} for fracture detection, while the second stage uses a novel multi-head classification network to characterize detected fractures across multiple clinically relevant dimensions.

\item We propose a hyperbolic multi-modal learning framework to integrate imaging and clinical text data. By projecting visual and textual features into hyperbolic space, our model captures hierarchical relationships in fracture classification, enhancing similarity computations and improving performance across multiple fracture attributes.

\end{itemize}

\begin{figure*}[!t]
    \centering
    \includegraphics[width=\textwidth,height=4cm]{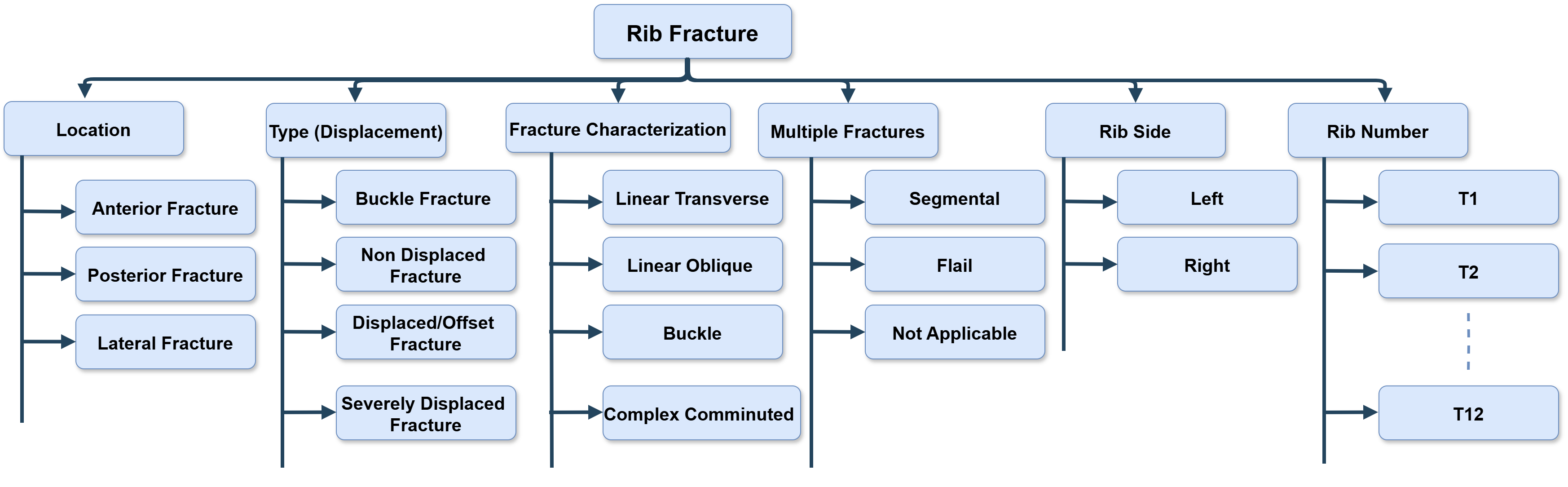}
    \caption{Flowchart illustrating the classification of rib fractures based on various parameters, including location, type (displacement), fracture characterization, multiple fractures, rib side, and rib number.}
    \label{fig:flowchart}
\end{figure*}

\section{Methodology}
This section describes our annotation protocol for rib fractures and the proposed DL model for fine-grained rib fracture analysis.

\subsection{Annotation Framework}

We establish a structured annotation framework based on established clinical classifications. This system incorporates anatomical parameters and the displacement taxonomy developed by the Chest Wall Injury Society (CWIS) consensus, categorizing fractures as undisplaced, offset, or displaced ~\cite{edwards2020taxonomy}. As shown in Figure \ref{fig:flowchart}, each rib fracture is annotated across six key dimensions: Location, Type (Displacement), Fracture Characterization, Multiple Fractures, Rib Side, and Rib Number. Each fracture in the AirRib dataset was manually annotated by two radiologists with oversight from senior trauma radiologists to resolve discrepancies. Fractures were delineated as volumetric regions of interest (VOI) on axial images. Each fracture was assigned a unique serial number and documented with rib side (left/right), rib number (T1–T12), location (anterior, lateral, posterior), displacement severity, and morphology. The contribution of fractures to flail or segmental injuries was also recorded. Each scan was assigned a randomized AIRib ID, linking CT volumes to segmentation masks and structured worksheets. A total of 1,014 rib fractures were annotated. For the public dataset, we randomly selected 50 CT scans from the RibFrac dataset~\cite{ribfracchallenge2024} and performed detailed labeling for each visualized traumatic rib fracture. Each fracture was labeled separately and assigned a unique serial number. The labeling process is similar to that followed for the AiRib dataset. A total of 473 rib fractures were annotated for the RibFrac dataset.
\begin{figure*}[!t]
    \centering
    \includegraphics[width=\textwidth,height=6cm]{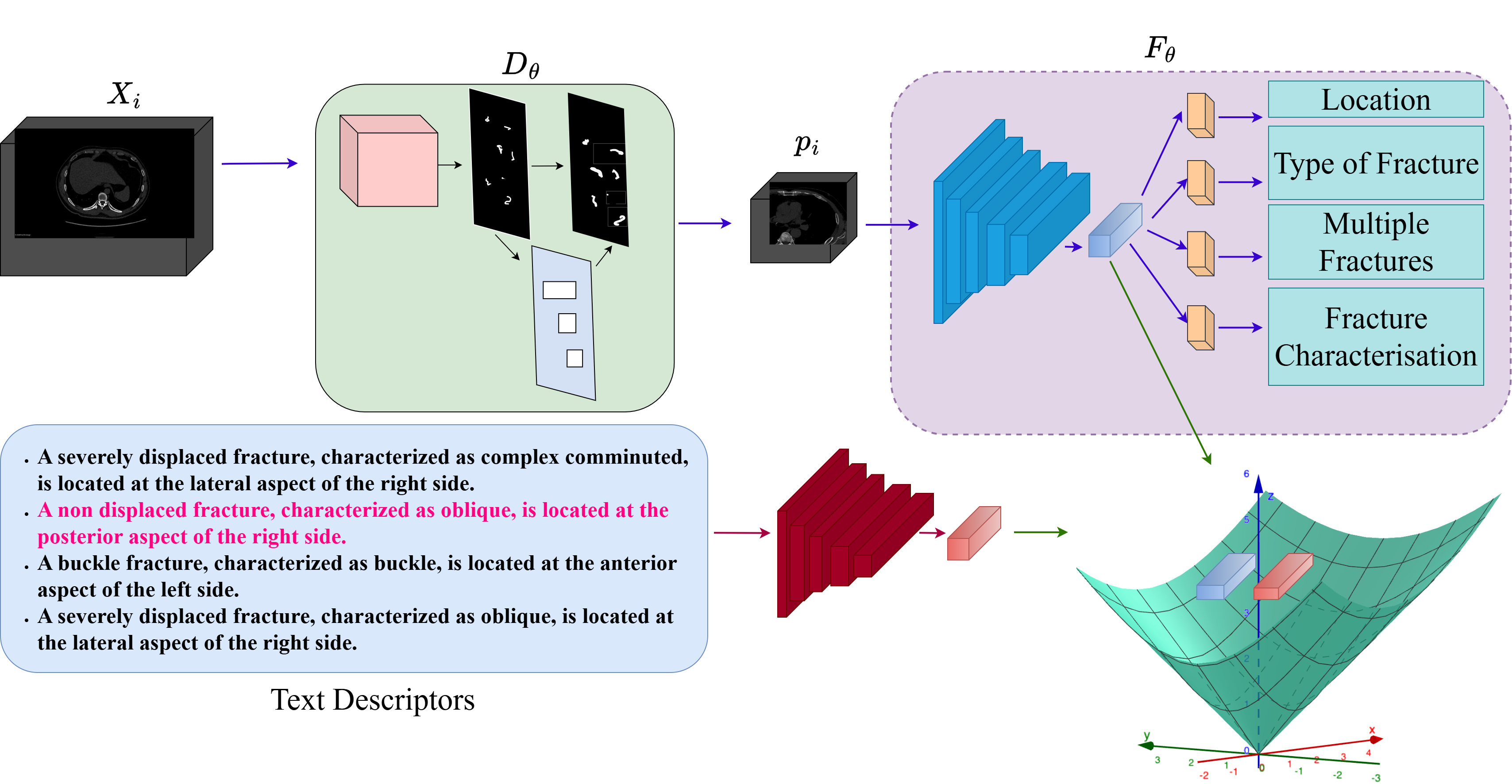}
    \caption{Overview of our proposed two-stage framework for fine-grained rib fracture analysis. The input CT scan ($X_i$) is processed by the detection model ($D_\theta$) to locate potential fracture patches ($p_i$). These patches are classified ($F_\theta$) for four distinct characteristics: Location, Type of Fracture, Multiple Fractures, and Fracture Characterization. Radiological descriptions are mapped alongside visual features to enable fine-grained classification.}
    \label{fig:model}
\end{figure*}

\subsection{Proposed Model}
As shown in Figure \ref{fig:model}, we design a fine-grained rib fracture analysis framework incorporating hyperbolic embeddings and multi-modal learning. Given a dataset $\mathcal{D} = {X_1, X_2, \ldots, X_N}$ consisting of $N$ patients, where each patient volume $X_i$ contains CT scans $X_i \in {R}^{H \times W \times D}$ and corresponding fracture annotations $Y_i = {y_1, y_2, ..., y_M}$, with $y_j$ representing individual fracture instances and their associated parameters. Our two-stage pipeline consists of a detection module, followed by a novel hyperbolic multi-head classification network. The framework is trained on our comprehensive annotation scheme, where clinical experts have labeled each fracture with detailed attributes that directly influence patient outcomes. This fine-grained approach not only provides detailed fracture characterization but also enables the automatic calculation of RibScore for objective severity assessment and treatment planning.

\subsubsection{Fracture Detection Module}

Our rib fracture detection module employs a two-stage object detection framework $D_{\theta}$ based on Faster R-CNN~\cite{ren2015faster}. The network incorporates a Region Proposal Network that generates potential fracture locations using customized anchor boxes at five scales (4, 8, 16, 32, and 64 pixels) and three aspect ratios (0.5, 1.0, and 2.0), followed by an ROI head for classification and bounding box refinement. Input CT slices $X_i$ undergo windowing to [-200, 1000] HU, normalization to [0, 1], and channel replication before processing. To address the volumetric nature of fractures despite using a 2D detection approach, we implement a post-processing fracture tracking algorithm that connects detections across adjacent slices. Detected bounding boxes are tracked between slices using spatial overlap and center-point proximity metrics.


Our temporal coherence algorithm ensures spatial consistency by grouping detections into coherent fracture tracks using an IoU threshold of 0.1\% between adjacent slices. Only tracks spanning at least four consecutive slices are retained to filter out spurious detections and ensure robust localization. Each validated track is represented by a 3D bounding box, from which standardized patches $\mathcal{\mathcal{P}} = {p_1, p_2, \ldots, p_t}$ of size $64\times64\times32$ voxels are extracted, capturing both the fracture site and surrounding anatomy. Zero-padding is applied for edge cases near volume boundaries, and identity affine transformations preserve spatial integrity, ensuring consistent input dimensions. After detection, volumetric patches (64×64×32) are extracted from the original CT and label volumes, centered on each detection’s geometric center, with zero-padding for boundary cases. These patches, along with metadata, are used as input to the fine-grained classification module.

\subsubsection{Fine Grained Classification Module}

For the fine-grained classification, we propose a hyperbolic multi-modal learning framework $F_{\theta}$ that jointly leverages CT imaging data and clinical text descriptions. The architecture comprises of three key components: a 3D image encoder with multi-task classification heads, a clinical text embedder with domain-specific language understanding, and a hyperbolic projection module that establishes semantic alignment between imaging features and textual descriptions in non-Euclidean space \ref{fig:model}.

CT patches ($p_i$) generated by the detection module are processed through a 3D ResNet-50 ~\cite{he2015deep} encoder from a U-Net ~\cite{ronneberger2015u} architecture, generating hierarchical feature maps. We extract final-layer features through adaptive 3D average pooling, producing image embeddings $\mathbf{v} \in {R}^{2048}$. Four parallel classification heads (location, fracture type, multiplicity, and characterization) process these features through linear layers with weighted cross-entropy loss to address class imbalance:


\begin{equation}
\mathcal{L}_{cls} = \sum_{h \in \mathcal{H}} \mathcal{L}_{CE}(\mathbf{W}_h\mathbf{v}, y_h, \omega_h)
\end{equation}

\noindent where \( \mathcal{H} = \{\text{head}_1, \text{head}_2, \text{head}_3, \text{head}_4\} \), and \( \omega_h \in \mathbb{R}^{n_h} \) represents the vector of class weights for head \( h \) with \( n_h \) classes.




Radiological descriptions are encoded using a pretrained ClinicalBERT ~\cite{huang2019clinicalbert} model fine-tuned on fracture reports. For a text description $T$, we extract the CLS token embedding $\mathbf{t} \in {R}^{1024}$ from the final transformer layer, capturing the global semantic context. Now to model the hierarchical nature of anatomical structures and fracture patterns, we project both modalities to a hyperboloid manifold using exponential mapping:

\begin{equation}
\text{Exp}_0(\mathbf{x}) = \sinh\left(\sqrt{|\kappa|} \|\mathbf{x}\|\right) \frac{\mathbf{x}}{\sqrt{|\kappa|}\|\mathbf{x}\|}
\end{equation}

\noindent where $\kappa < 0$ is the learnable curvature parameter. Image ($\mathbf{v}'$) and text ($\mathbf{t}'$) embeddings are projected through separate geodesic projection heads:

\begin{equation}
\mathbf{v}' = \text{Proj}_{\text{img}}(\mathbf{v}); \quad \mathbf{t}' = \text{Proj}_{\text{txt}}(\mathbf{t})
\end{equation}

We employ a multi-objective loss combining, Geodesic Contrastive Loss ~\cite{desai2023hyperbolic} and Entailment Cone Loss ~\cite{desai2023hyperbolic}. The Geodesic Contrastive Loss  Minimizes hyperbolic distance between matched image-text pairs, while the Entailment Cone Loss Constraints text embeddings to reside in the hyperbolic cone of corresponding image features. For rib number and side determination, we implement a computer vision-based approach. This method relies on spatial positioning within the full CT volume. Side determination employs a threshold-based technique that compares the patch's center position to the midline of the thorax, classifying fractures as left or right based on their relative position. For specific ribs like T1, we combine positional criteria with structural analysis to achieve accurate identification. During inference, we implement a consensus mechanism: if $\geq$2 classification heads predict ``no fracture'', all heads are constrained to negative predictions. This hierarchical decision process mimics clinical reasoning by prioritizing specificity.

\section{Experiment \& Results}
\subsection{Dataset}

We utilize two datasets: our proprietary AirRib dataset and the publicly available RibFrac dataset~\cite{ribfracchallenge2024}. The AirRib dataset has 103 CT scans from 137 trauma patients from a government hospital, selected through strict inclusion criteria: complete thoracic coverage showing all 12 ribs and the presence of at least one fracture. Exclusion criteria removed scans with motion artifacts, non-traumatic bone pathologies, prior surgical interventions, incomplete sections, or missing thin-slice reconstructions. Imaging was performed on three Siemens Healthineers scanners (Somatom Sensation 40, Somatom Definition AS, and Somatom Definition Edge) using standard trauma protocols. Data retrieval complied with HIPAA guidelines, and the study was approved by the Institutional Ethics Committee, \footnote{IECPG-155/24.02.2021}, with informed consent waived due to its retrospective design.

For the classification module, we generate textual descriptions of each fracture using the assigned labels, e.g if a fracture is labeled as occurring on the right rib side in the lateral region, as a non-displaced fracture oblique fracture, the generated description would be: ``The fracture is located on the right side of the rib, specifically in the lateral region. It is a non-displaced fracture with an oblique pattern.'' For the detection model, we use the entire RibFrac dataset except for the 50 samples, which are split into 40 for training and 10 for testing. We fine-tune the detection model using these 40 training samples and evaluate the complete rib fracture analysis framework, from detection to classification, using the remaining 10 samples. Similarly, for the AirRib dataset, we use 84 samples for training and 19 for testing, fine-tuning the detection model with the 84 training samples and evaluating it on the 19 test samples.


\subsection{Implementation Details}

For the detection module, we employ SGD~\cite{ruder2016overview} (learning rate=$1 \times 10^{-4}$, momentum=0.9, weight decay=$5 \times 10^{-4}$) with a ResNeXt-101-32$\times$8d~\cite{xie2017aggregated} feature pyramid~\cite{lin2017feature} backbone. We filtered detections using a 0.5 confidence threshold and removed predictions larger than $80 \times 80$ pixels to reduce false positives. The classification model was trained using Adam (learning rate=$1 \times 10^{-4}$, weight decay=$1 \times 10^{-5}$) with 5-epoch linear warmup. Other hyperparameters include curvature $\kappa=1.0$, entailment weight $\lambda=0.2$, and temperature $\tau=0.07$. Data augmentation employed random 3D translations ($\pm$10px), rotations ($\pm$10$^{\circ}$), and intensity windowing ([-200, 1000] HU). We benchmark against rib fracture classification models (UCI and CS ~\cite{yang2024deep}) and state-of-the-art bone fracture classification (CT ~\cite{zhu2023cross}).

\begin{table*}[!t]
\setlength{\tabcolsep}{2pt}

\centering
\caption{Performance comparison of different models on AirRib and RibFrac datasets showing Accuracy, Precision, and Recall metrics per model}
\begin{tabular}{l|c|cccc|cccc|cccc}
\hline
\multirow{2}{*}{Dataset} & \multirow{2}{*}{CLS Head} & \multicolumn{4}{c|}{Accuracy} & \multicolumn{4}{c|}{Recall} & \multicolumn{4}{c}{Precision} \\
\cline{3-14}
 &  & CT & CS & UCI & Ours & CT & CS & UCI & Ours & CT & CS & UCI & Ours \\
\hline
\multirow{5}{*}{AirRib} & Location & 0.40 & 0.42 & 0.40 & \textbf{0.44} & 0.59 & 0.56 & 0.58 & \textbf{0.61} & 0.45 & 0.43 & 0.46 & \textbf{0.42}\\
 & Type of frac. & 0.32 & 0.34 & 0.30 & \textbf{0.34} & 0.48 & 0.46 & 0.50 & \textbf{0.49} & 0.30 & 0.31 & 0.32 & \textbf{0.36}\\
 & Multiple frac. & 0.31 & 0.39 & 0.32 & \textbf{0.35} & 0.31 & 0.35 & 0.35 & \textbf{0.41} & 0.36 & 0.38 & 0.38 & \textbf{0.41}\\
 & Frac. Charac. & 0.40 & 0.34 & 0.36 & \textbf{0.39} & 0.19 & 0.65 & 0.43 & \textbf{0.61} & 0.25 & 0.31 & 0.30 & \textbf{0.29}\\
\cline{2-14}
 & \textbf{Average} & 0.36 & 0.37 & 0.35 & \textbf{0.38} & 0.39 & 0.50 & 0.47 & \textbf{0.53} & 0.34 & 0.36 & 0.36 & \textbf{0.37}\\
\hline
\multirow{5}{*}{RibFrac} & Location & 0.48 & 0.50 & 0.40 & \textbf{0.53} & 0.50 & 0.52 & 0.52 & \textbf{0.61} & 0.42 & 0.43 & 0.45 & \textbf{0.47}\\
 & Type of frac. & 0.35 & 0.36 & 0.29 & \textbf{0.45} & 0.32 & 0.28 & 0.29 & \textbf{0.46} & 0.23 & 0.26 & 0.24 & \textbf{0.37}\\
 & Multiple frac. & 0.45 & 0.45 & 0.35 & \textbf{0.45} & 0.33 & 0.44 & 0.27 & \textbf{0.43} & 0.32 & 0.33 & 0.28 & \textbf{0.35}\\
 & Frac. Charac. & 0.37 & 0.37 & 0.30 & \textbf{0.43} & 0.36 & 0.37 & 0.29 & \textbf{0.39} & 0.26 & 0.35 & 0.24 & \textbf{0.30}\\
\cline{2-14}
 & \textbf{Average} & 0.41 & 0.42 & 0.34 & \textbf{0.46} & 0.37 & 0.40 & 0.34 & \textbf{0.47} & 0.30 & 0.34 & 0.30 & \textbf{0.37}\\
\hline
\end{tabular}
\label{tab:classification_results}
\end{table*}
\subsection{Results}

Table~\ref{tab:classification_results} demonstrates our model's superior performance against three baselines (CT, CS, UCI) across both datasets. We achieved higher average recall (AirRib: 0.53, RibFrac: 0.47) compared to CT (0.39, 0.37) and CS (0.50, 0.40). Notable improvements include Multiple fractures (0.41 vs.\ 0.35 CS) and Fracture Characterization (0.61 vs.\ 0.19 CT) on AirRib, and Location (0.61 vs.\ 0.52 CS) and Type of fracture (0.46 vs.\ 0.32 CT, 0.28 CS) on RibFrac. We also validate our model's clinical relevance by evaluating its ability to accurately derive RibScore components from predicted fracture characteristics. For example, patient ID 84 has a fracture in the T1 rib and two segmental fractures, yielding a ground truth RibScore of 1. Our model correctly identifies the T1 and the two segmental fractures, resulting in a matching predicted RibScore of 1.
In a more complex case, patient ID 88 presents with a T1 fracture, flail chest, multiple severely displaced fractures, and fractures across all three anatomical areas, but no bilateral fractures. This configuration corresponds to a ground truth RibScore of 5. Our model successfully identifies all these critical characteristics, resulting an accurate predicted RibScore of 5. These examples demonstrate our framework's ability to capture clinically meaningful fracture patterns directly translating to accurate severity assessment, supporting its potential utility in clinical decision-making.

Our ablation study (Table~\ref{tab:ab}) confirms the significant impact of textual descriptors. Incorporating these descriptors increased recall from 0.48 to 0.53 on AirRib (with Fracture Characterization improving from 0.41 to 0.61) and from 0.405 to 0.47 on RibFrac (with Location improving from 0.55 to 0.61 and Type of fracture from 0.37 to 0.46). These results demonstrate that textual descriptors provide valuable semantic context that complements visual features, particularly for ambiguous imaging features.

\begin{table*}[!t]
\setlength{\tabcolsep}{8pt}
\centering
\caption{Ablation results in terms of accuracy (Acc), precision (Pre) and recall (Rec) without textual descriptors using Cross Entropy Loss.}
\begin{tabular}{l|ccc|ccc}
\hline
\multirow{2}{*}{CLS Head} & \multicolumn{3}{c|}{AirRib} & \multicolumn{3}{c}{RibFrac} \\
 & Acc & Rec & Pre & Acc & Rec & Pre \\
\hline
Location & 0.43 & 0.59 & 0.42 & 0.42 & 0.55 & 0.49 \\
Type of frac. & 0.30 & 0.55 & 0.34 & 0.35 & 0.37 & 0.34 \\
Multiple frac. & 0.32 & 0.36 & 0.38 & 0.45 & 0.29 & 0.31 \\
Frac. Charac. & 0.42 & 0.41 & 0.27 & 0.35 & 0.40 & 0.38 \\
\hline
\textbf{Average} & 0.37 & 0.48 & 0.35 & 0.39 & 0.40 & 0.38 \\
\hline
\end{tabular}
\label{tab:ab}
\end{table*}

\section{Conclusion}
This work introduces a novel method for rib fracture analysis using fine-grained annotation and hyperbolic embedding-based classification. By developing a detailed annotation framework for clinically relevant fracture attributes and a hyperbolic multi-modal learning architecture, we show significant performance improvements over existing approaches. Our two-stage pipeline enables precise classification across location, displacement severity, and morphology. Combining radiological descriptions with visual features provides complementary insights. This approach advances automated rib fracture analysis and enhances RibScore calculation and treatment planning. Future work will expand this framework to larger multi-institutional datasets and explore its application to other anatomical structures with hierarchical relationships.

%
%
%
 \bibliographystyle{splncs04}
 \bibliography{mybib}

\begin{thebibliography}{10}
\providecommand{\url}[1]{\texttt{#1}}
\providecommand{\urlprefix}{URL }
\providecommand{\doi}[1]{https://doi.org/#1}

\bibitem{cao2023robust}
Cao, Z., Xu, L., Chen, D.Z., Gao, H., Wu, J.: A robust shape-aware rib fracture detection and segmentation framework with contrastive learning. IEEE Transactions on Multimedia  \textbf{25},  1584--1591 (2023)

\bibitem{castro2024ribfracturesys}
Castro-Zunti, R., Li, K., Vardhan, A., Choi, Y., Jin, G.Y., Ko, S.b.: Ribfracturesys: A gem in the face of acute rib fracture diagnoses. Computerized Medical Imaging and Graphics  \textbf{117},  102429 (2024)

\bibitem{chapman2016ribscore}
Chapman, B.C., Herbert, B., Rodil, M., Salotto, J., Stovall, R.T., Biffl, W., Johnson, J., Burlew, C.C., Barnett, C., Fox, C., et~al.: Ribscore: a novel radiographic score based on fracture pattern that predicts pneumonia, respiratory failure, and tracheostomy. Journal of Trauma and Acute Care Surgery  \textbf{80}(1),  95--101 (2016)

\bibitem{desai2023hyperbolic}
Desai, K., Nickel, M., Rajpurohit, T., Johnson, J., Vedantam, S.R.: Hyperbolic image-text representations. In: International Conference on Machine Learning. pp. 7694--7731. PMLR (2023)

\bibitem{edwards2020taxonomy}
Edwards, J.G., Clarke, P., Pieracci, F.M., Bemelman, M., Black, E.A., Doben, A., Gasparri, M., Gross, R., Jun, W., Long, W.B., et~al.: Taxonomy of multiple rib fractures: results of the chest wall injury society international consensus survey. Journal of trauma and acute care surgery  \textbf{88}(2),  e40--e45 (2020)

\bibitem{franssen2024treatment}
Franssen, A.J., Daemen, J.H., Luyten, J.A., Meesters, B., Pijnenburg, A.M., Reisinger, K.W., van Vugt, R., Hulsew{\'e}, K.W., Vissers, Y.L., de~Loos, E.R.: Treatment of traumatic rib fractures: an overview of current evidence and future perspectives. Journal of Thoracic Disease  \textbf{16}(8), ~5399 (2024)

\bibitem{he2015deep}
He, K., Zhang, X., Ren, S., Sun, J.: Deep residual learning for image recognition. corr abs/1512.03385 (2015) (2015)

\bibitem{huang2019clinicalbert}
Huang, K., Altosaar, J., Ranganath, R.: Clinicalbert: Modeling clinical notes and predicting hospital readmission. arXiv preprint arXiv:1904.05342  (2019)

\bibitem{jin2020deep}
Jin, L., Yang, J., Kuang, K., Ni, B., Gao, Y., Sun, Y., Gao, P., Ma, W., Tan, M., Kang, H., et~al.: Deep-learning-assisted detection and segmentation of rib fractures from ct scans: Development and validation of fracnet. EBioMedicine  \textbf{62} (2020)

\bibitem{khrulkov2020hyperbolic}
Khrulkov, V., Mirvakhabova, L., Ustinova, E., Oseledets, I., Lempitsky, V.: Hyperbolic image embeddings. In: Proceedings of the IEEE/CVF conference on computer vision and pattern recognition. pp. 6418--6428 (2020)

\bibitem{li2023automatic}
Li, N., Wu, Z., Jiang, C., Sun, L., Li, B., Guo, J., Liu, F., Zhou, Z., Qin, H., Tan, W., et~al.: An automatic fresh rib fracture detection and positioning system using deep learning. The British Journal of Radiology  \textbf{96}(1146),  20221006 (2023)

\bibitem{lin2017feature}
Lin, T.Y., Doll{\'a}r, P., Girshick, R., He, K., Hariharan, B., Belongie, S.: Feature pyramid networks for object detection. In: Proceedings of the IEEE conference on computer vision and pattern recognition. pp. 2117--2125 (2017)

\bibitem{liu2021multi}
Liu, J., Cui, Z., Sun, Y., Jiang, C., Chen, Z., Yang, H., Zhang, Y., Wu, D., Shen, D.: Multi-scale segmentation network for rib fracture classification from ct images. In: Machine Learning in Medical Imaging: 12th International Workshop, MLMI 2021, Held in Conjunction with MICCAI 2021, Strasbourg, France, September 27, 2021, Proceedings 12. pp. 546--554. Springer (2021)

\bibitem{ren2015faster}
Ren, S., He, K., Girshick, R., Sun, J.: Faster r-cnn: Towards real-time object detection with region proposal networks. Advances in neural information processing systems  \textbf{28} (2015)

\bibitem{ren2016faster}
Ren, S., He, K., Girshick, R., Sun, J.: Faster r-cnn: Towards real-time object detection with region proposal networks. IEEE transactions on pattern analysis and machine intelligence  \textbf{39}(6),  1137--1149 (2016)

\bibitem{ringl2015ribs}
Ringl, H., Lazar, M., T{\"o}pker, M., Woitek, R., Prosch, H., Asenbaum, U., Balassy, C., Toth, D., Weber, M., Hajdu, S., et~al.: The ribs unfolded-a ct visualization algorithm for fast detection of rib fractures: effect on sensitivity and specificity in trauma patients. European radiology  \textbf{25},  1865--1874 (2015)

\bibitem{ronneberger2015u}
Ronneberger, O., Fischer, P., Brox, T.: U-net: Convolutional networks for biomedical image segmentation. In: Medical image computing and computer-assisted intervention--MICCAI 2015: 18th international conference, Munich, Germany, October 5-9, 2015, proceedings, part III 18. pp. 234--241. Springer (2015)

\bibitem{ruder2016overview}
Ruder, S.: An overview of gradient descent optimization algorithms. arXiv preprint arXiv:1609.04747  (2016)

\bibitem{wu2021development}
Wu, M., Chai, Z., Qian, G., Lin, H., Wang, Q., Wang, L., Chen, H.: Development and evaluation of a deep learning algorithm for rib segmentation and fracture detection from multicenter chest ct images. Radiology: Artificial Intelligence  \textbf{3}(5),  e200248 (2021)

\bibitem{xie2017aggregated}
Xie, S., Girshick, R., Doll{\'a}r, P., Tu, Z., He, K.: Aggregated residual transformations for deep neural networks. In: Proceedings of the IEEE conference on computer vision and pattern recognition. pp. 1492--1500 (2017)

\bibitem{ribfracchallenge2024}
Yang, J., Shi, R., Jin, L., Huang, X., Kuang, K., Wei, D., Gu, S., Liu, J., Liu, P., Chai, Z., Xiao, Y., Chen, H., Xu, L., Du, B., Yan, X., Tang, H., Alessio, A., Holste, G., Zhang, J., Wang, X., He, J., Che, L., Pfister, H., Li, M., Ni, B.: Deep rib fracture instance segmentation and classification from ct on the ribfrac challenge. arXiv Preprint  (2024)

\bibitem{yang2024deep}
Yang, J., Shi, R., Jin, L., Huang, X., Kuang, K., Wei, D., Gu, S., Liu, J., Liu, P., Chai, Z., et~al.: Deep rib fracture instance segmentation and classification from ct on the ribfrac challenge. arXiv preprint arXiv:2402.09372  (2024)

\bibitem{zhou2021automatic}
Zhou, Q.Q., Tang, W., Wang, J., Hu, Z.C., Xia, Z.Y., Zhang, R., Fan, X., Yong, W., Yin, X., Zhang, B., et~al.: Automatic detection and classification of rib fractures based on patients’ ct images and clinical information via convolutional neural network. European Radiology  \textbf{31},  3815--3825 (2021)

\bibitem{zhu2023cross}
Zhu, Z., Chen, Q., Yu, L., Wang, L., Zhang, D., Magnier, B., Wang, L.: Cross-view deformable transformer for non-displaced hip fracture classification from frontal-lateral x-ray pair. In: International Conference on Medical Image Computing and Computer-Assisted Intervention. pp. 444--453. Springer (2023)

\end{thebibliography}

\end{document}